\documentclass[lettersize,journal]{IEEEtran}
\usepackage[ruled]{algorithm2e}
\usepackage{array}
\usepackage[caption=false,font=normalsize,labelfont=sf,textfont=sf]{subfig}
\usepackage{textcomp}
\usepackage{amsmath,amssymb,amsfonts}
\usepackage{graphicx}
\usepackage{wrapfig}
\DeclareMathOperator{\E}{\mathbb{E}}
\usepackage[caption=false,font=normalsize,labelfont=sf,textfont=sf]{subfig}
\usepackage{stfloats}
\usepackage{url}
\usepackage{verbatim}
\usepackage{color}
\usepackage{cite}
\usepackage{booktabs}
\usepackage{multirow}
\usepackage{makecell}
\usepackage{amsmath}
\usepackage{amssymb}
\usepackage[frak=mma]{mathalfa}
\usepackage{bbm}
\usepackage[mathscr]{euscript}
\usepackage{stfloats}
\usepackage{verbatim}
\usepackage{cite}
\usepackage{caption}
\usepackage{bbm}
\usepackage{comment}

\usepackage[pagebackref,breaklinks,colorlinks]{hyperref}
\newcommand{\twopartdef}[4]
{
	\left\{
		\begin{array}{ll}
			#1 & \mbox{if } #2 \\
			#3 & \mbox{if } #4
		\end{array}
	\right.
}

\begin{document}

\title{AURA : Automatic Mask Generator using \\ Randomized Input Sampling for Object Removal}

\author{Changsuk Oh and H. Jin Kim
\thanks{Changsuk Oh and H. Jin Kim with the Department of Aerospace Engineering, Seoul National University, Seoul, Korea (email: santgo@snu.ac.kr and hjinkim@snu.ac.kr)}
}

\markboth{Journal of \LaTeX\ Class Files,~Vol.~14, No.~8, August~2021}%
{Shell \MakeLowercase{\textit{et al.}}: A Sample Article Using IEEEtran.cls for IEEE Journals}

\maketitle

\begin{abstract}
The objective of the image inpainting task is to fill missing regions of an image in a visually plausible way. Recently, deep learning-based image inpainting networks have generated outstanding results, and some utilize their models as object removers by masking unwanted objects in an image. Previous works try to enhance the quality of object removal results only through improvements in the inpainting performance of their networks and pay less attention to the importance of the input mask. In this paper, we focus on generating the input mask to better remove objects using off-the-shelf image inpainting networks. We propose an automatic mask generator inspired by an explainable AI (XAI) method, whose output can better remove objects than masks obtained using a segmentation network. The proposed method generates an importance map using randomly sampled input masks and estimated scores of the completed images obtained from the random masks. The output mask is selected by a judge module among the candidate masks which are generated from the importance map. Experiments confirm that input masks generated by our method can better remove objects than those generated from a segmentation network.
\end{abstract}

\section{Introduction}

Object removal aims to erase instances in an image while preserving the overall appearance. As it is challenging to capture an image in a controlled environment where only desired foreground objects appear, object removal is a highly sought-after technology in the real world. Recently, \cite{liu2018image, yi2020contextual, cao2021learning, rombach2022high, zeng2020high, suvorov2022resolution, zeng2022aggregated} generates high-quality object removal results using their deep learning-based inpainting networks.
The object removal quality is heavily affected by an input mask shape and the performance of an image inpainting network. Although the output quality is heavily influenced by the two factors, previous works \cite{liu2018image, yi2020contextual, cao2021learning, rombach2022high, zeng2020high, suvorov2022resolution, zeng2022aggregated} focus only on enhancing the performance of their inpainting network to acquire high-quality outputs. The above studies utilize semantic segmentation maps or human-made patches to make input masks, but there is \textit{no} discussion on which mask can better erase removal targets using their networks. 

\begin{figure}[t!]
\centering
\includegraphics[width=1.0\linewidth]{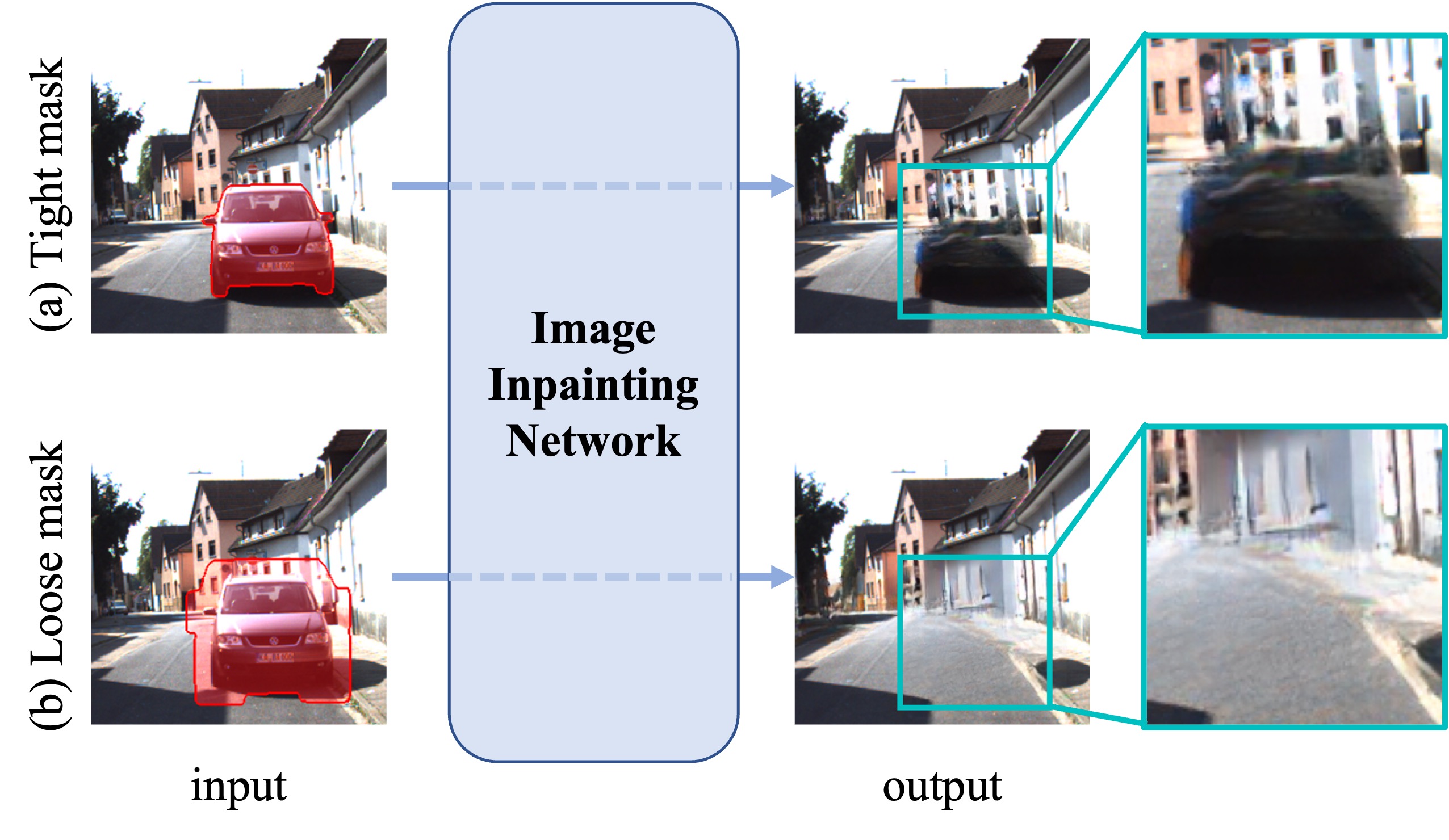} 
\caption{Object removal results obtained using the tight and loose masks. \cite{suvorov2022resolution} is utilized for inpainting.}
\label{fig:thumbnail}
\end{figure}

In this paper, we focus on the shape of input masks to obtain high-quality object removal results. When the input mask fails to cover all pixels of a removal target, the afterimage of the target may exist in the object removal result. We can find such afterimages in the object removal results in Fig. \ref{fig:thumbnail} (a). On the other hand, if the input mask is too large compared to a removal target, it covers surrounding pixels of the removal targets. In such cases, an image inpainting network may encounter difficulties in plausibly restoring the surrounding areas of the removal target, as shown in Fig. \ref{fig:thumbnail} (b). Therefore, the input mask should completely cover removal targets while not masking key areas for completing missing areas to obtain satisfactory object removal results.

To perform object removal, it is necessary to specify removal targets in an image. Excluding the use of human-made patches for masking, a removal target is needed to be specified by clicking a pixel of the target or by providing verbal instructions in human language. In such cases, a user needs a segmentation module to mask removal targets. In this paper, we refer to the mask that covers the pixels corresponding to the removal targets, identified using a segmentation network, as a target segmentation mask. We use target segmentation masks and their variants (enlarged segmentation masks) as baselines. Our goal is to generate input masks that can better remove objects than the baseline masks.

To generate the better input mask, we propose a new mask generation method called AUtomatic mask generator using RAndomized input sampling (AURA). This work is inspired by the explainable AI (XAI) method \cite{petsiuk2018rise}, which tries to understand a black-box classification network’s prediction. The proposed mask generator randomly samples input masks and gets restored images using them. Then, AURA generates an importance map.  An importance of a pixel represents the expected quantitative score of the completed image when the pixel is masked. We use a weighted linear combination of random masks to generate an importance map, where the quantitative evaluations of the masks are used as scaling factors. We design the judge module to evaluate the quality of the outputs by quantifying 1) how much the afterimage of the target objects remains and 2) whether the backgrounds have been well restored. AURA generates candidate masks by masking pixels that exceed predefined percentile values of the importance map, and the judge module figures out the best mask among them. In this way, we can make the input masks for better removal. We test our method on the COCO \cite{lin2014microsoft} and RORD \cite{Sagong_2022_BMVC} datasets. In this paper, we evaluate an object remover highly if it preserves the original image well while leaving no visual features of removal targets.

This paper has the following contributions:
\begin{itemize}
    \item This is the first paper that focuses on the importance of input mask to better remove objects in an image using the image inpainting network.
    \item We propose AURA, an automatic mask generator that produces a mask which can better remove objects than a target segmentation mask and its variants (enlarged segmentation masks).
    \item Experiments on the large image datasets demonstrate that the masks generated by the proposed method can better remove objects than those obtained by a semantic segmentation network. 
\end{itemize}

\section{Related Work}

\subsection{Image Inpainting}
Before the deep learning era, image inpainting methods utilize visual features around the missing areas to complete them. \cite{bertalmio2000image, 1217265, 935036} introduce ways to fill holes by considering continuity of visual features with surroundings, such as pixel values or gradient orientation. \cite{6960838, 5685527, kwatra2005texture, efros2023image} propose various approaches for finding the most suitable patch in unmasked areas to fill a masked region. After the advent of the deep learning era, various deep learning techniques have been utilized for image inpainting. Various approaches using partial convolution \cite{liu2018image}, Fourier concolution \cite{suvorov2022resolution}, gated convolution \cite{yu2019free}, recurrent reasoning \cite{li2020recurrent, lugmayr2022repaint}, adversarial loss \cite{jam2021r, wadhwa2021hyperrealistic, cao2021learning, iizuka2017globally, quan2022image, zeng2021cr, liu2021pd, Yan_2018_ECCV, zeng2020high, yi2020contextual, suvorov2022resolution, zeng2022aggregated, Wang_2022_CVPR}, attention propagation module \cite{liao2021image}, latent diffusion model \cite{rombach2022high}, contextual attention \cite{yu2018generative}, and transformer \cite{li2022mat, dong2022incremental, liu2022reduce} are proposed for image inpainting. 

Previous works deploy their networks for various applications. \cite{Liu_2021_WACV, Niu_2023_ICCV, zhao2022visible}, and \cite{Yan_2022_WACV, 9423583, shao2021selective} utilize inpainting networks for watermark removal, and raindrop removal, respectively. An image inpainting network can also be utilized for object removal by masking unwanted objects in an image. \cite{liu2018image, suvorov2022resolution, zeng2022aggregated,cao2021learning, rombach2022high,yi2020contextual, zeng2020high, 9775085, 9751095} demonstrate object removal results obtained using their image inpainting network. In this paper, we focus on object removal and introduce a method for generating masks that can yield high-quality object removal results.

\subsection{Mask shapes for image inpainting networks}
The existing works utilize various sizes and shapes of masks to generate training data for image inpainting networks. Rectangular-shaped patches \cite{yu2018generative, iizuka2017globally, pathak2016context, Yan_2018_ECCV, Yang_2017_CVPR, wang2019laplacian, zhang2022gan}, irregular-shaped patches \cite{liu2018image, li2020recurrent, liu2021pd, Yu_2021_ICCV, li2020deepgin, wang2020vcnet, shin2020pepsi++, quan2022image, liu2022reduce, zeng2022aggregated}, and object-shaped patches \cite{rombach2022high, yi2020contextual, cao2021learning} are exploited to generate training data. 

However, the importance of input masks in generating high-quality inpainted images has not been sufficiently studied. Especially in the case of object removal, an input mask significantly affects the quality of the inpainted image. Specifically, a mask that tightly covers a removal target can minimize loss of background information. At the same time, unmasked pixels from the removal target can generate afterimage in the inpainted image. On the contrary, if a mask covers not only a removal target but also the surrounding area extensively, excessive loss of background information may result in less satisfactory background restoration. Therefore, to obtain decent object removal results, it is important to completely cover removal targets while preserving key surrounding information crucial for inpainting the masked regions. 

Even though an image inpainting network and input mask are essential elements for obtaining high-quality object removal results, previous works \cite{liu2018image, yi2020contextual, cao2021learning, rombach2022high, zeng2020high, suvorov2022resolution, zeng2022aggregated} have focused on achieving decent object removal results only through improvements in the performance of image inpainters. In this paper, we focus on obtaining an input mask that can better remove objects than masks generated from a semantic segmentation map.

\section{Method}

\begin{figure*}[t!]
\centering
\includegraphics[width=0.88\linewidth]{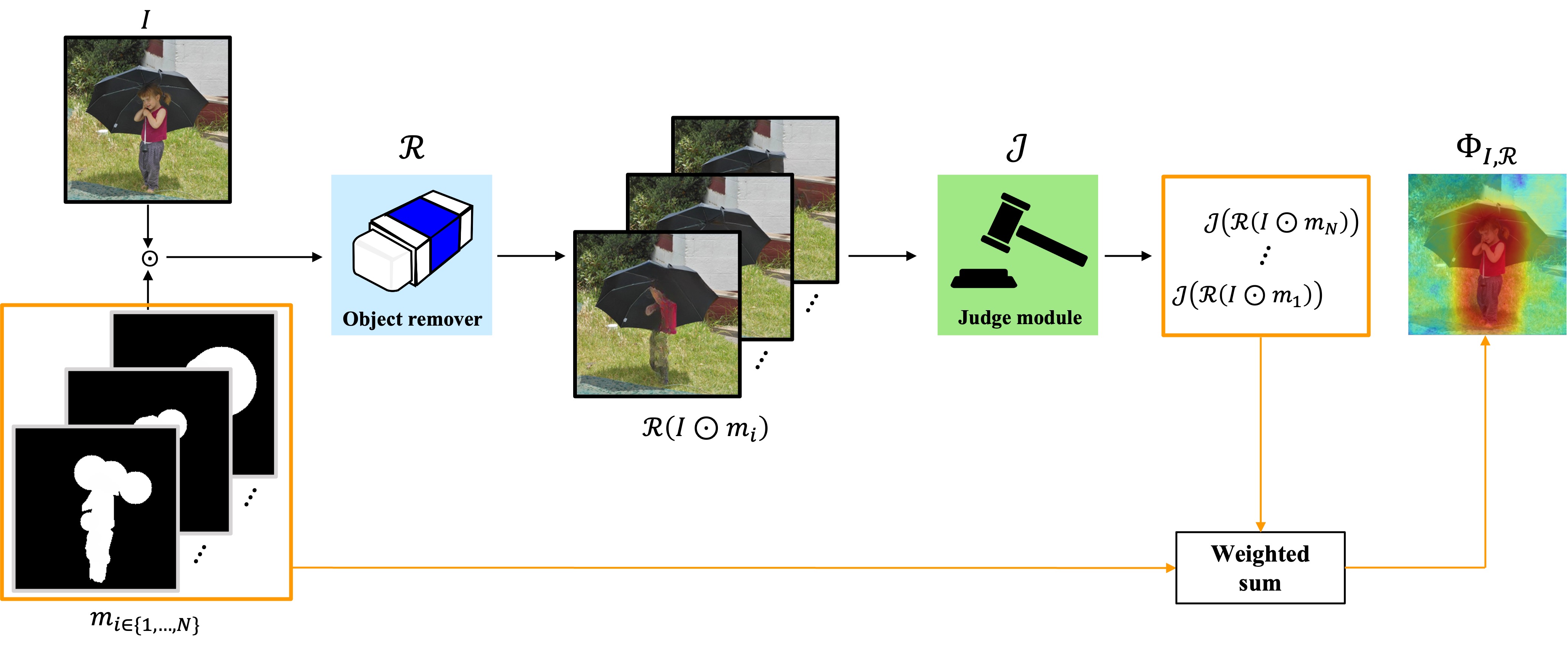} 
\caption{Flowchart of the importance map generation process.}
\label{fig:flowchart}
\end{figure*}

\subsection{Preliminary}
Randomized Input Sampling for Explanation (RISE) \cite{petsiuk2018rise} generates a saliency map which shows importance of each pixel for a black-box classification network’s inference. Unlike white-box networks, a user cannot access the parameters of a black-box model. Therefore, \cite{petsiuk2018rise} uses randomly sampled inputs (masks) and the outputs to understand a black-box model. Specifically, the authors define the importance of a pixel $x \in \Omega$ as the expected confidence score when the pixel is not masked, where $\Omega = \{1,...,H\} \times \{1, …, W\}$. Then, the importance of a pixel can be written as
\begin{align}
    S_{I,f} (x) = \E_{M} [ \ f(\ I \odot M\ ) \, |\,  M(x) = 1 \ ],
\end{align} 
where $f$ represents a black-box classifier that maps an image $I$ to a confidence score, and $ M : \Omega \mapsto \{ 0, 1 \}$ indicates a random mask with distribution $\mathcal{D}$. $\odot$ denotes the element-wise multiplication.

\subsection{Automatic Mask Generator using Random Input Sampling}
In this section, we introduce AUtomatic mask generator using RAndom input sampling (AURA), an automatic mask-generating method for object removal tasks. AURA makes an importance map using randomly sampled masks and completed images, which is depicted in Fig. \ref{fig:flowchart} and Algorithm \ref{algo:AURA}. Similar to the saliency map of \cite{petsiuk2018rise}, the importance map of AURA shows the expected quality of the completed image when each pixel is masked. In other words, an object remover can more plausibly erase objects when pixels with high importance are masked. The proposed method generates candidate masks using the importance map, and the judge module selects the mask that can best remove objects among them.

\textbf{Mask Generation }
We sample random masks $\{m_1, ... , m_N\}$ by adding  patches to a target segmentation map where only removal targets are labeled. Consequently, AURA can focus on figuring out which pixels should be further masked in the segmentation map rather than searching pixels among the entire image. We randomly sample the number, size, and location of additional patches, and details about the sampling process used for experiments are presented in Algorithm \ref{algo:mask}. 

\subsection{Judge module}
Utilizing a random mask $m_{i\in \{1, ..., N\}} $ and an image inpainting network (object remover) $\mathcal{R}$, we can obtain the completed image $\mathcal{R}(I \odot m_{i})$. Using the judge module, the proposed method quantitatively evaluates the judge score $\mathcal{J}$, which represents how well objects in the images are removed. The evaluation scores are used as weights for a linear combination of the masks to generate the importance map. The judge module evaluates how much 1) objects remain (Eq. \ref{eq:detect}), 2) the afterimage of the target objects remains (Eq. \ref{eq:afterimage}), and 3) the background is deformed (Eq. \ref{eq:background}). 

The 2D object detector \cite{redmon2018yolov3} is employed to find visual features of removal targets in the completed image. The judge module gives a low score when the objects are detected in the completed image. $\mathcal{J}_{detect}$ is calculated as follows:
\begin{align}
    \mathcal{J}_{detect} = - \frac{\mathcal{A}(\mathcal{S}(\mathcal{R}(I \odot m_i ))) }{\mathcal{A}(L)} 
    \label{eq:detect}
\end{align} 
Let $\mathcal{S} :\Omega \to \{0,1\}$ makes a binary map, which marks the detected areas as 1. $\mathcal{A}$ counts the number of pixels whose pixel value is 1 in a binary map. Pixels from removal targets are marked as 1 in the target segmentation map $L$. When the size of an object is small, even if it is detected, it does not significantly affect the overall score. To properly evaluate whether objects are removed well regardless of the objects' size, the size of the target segmentation mask is divided. 

$\mathcal{J}_{background}$ shows the similarity between the original image's background region and the completed image's corresponding region.  We convert the RGB intensities of  the target objects' pixels to zero. $\mathcal{J}_{background}$ is designed to give a low score if the background of the completed image is different from the original image's background. The judge module utilizes negative L2 to calculate the differences between the two manipulated images. We calculate $\mathcal{J}_{background}$ as follows:
\begin{align}
    \mathcal{J}_{background} = 
    - \frac{\mathcal{L}_{2} (I \odot (\mathbbm{1}_{I} - L), \mathcal{R} (  I \odot m_i )  \odot (\mathbbm{1}_{I} - L))}{ H \cdot W - \mathcal{A}(L)}.
    \label{eq:background}
\end{align}

We find that a mask that tightly covers an object leaves an afterimage of the object in the completed image, as shown in Fig. \ref{fig:thumbnail}. $\mathcal{J}_{afterimage}$ is designed to give a low score if there is an afterimage of the target objects. It calculates the perceptual difference between the completed images using target segmentation masks and query masks. Pixel values are converted to zeros except for pixels from the target objects to focus on regions of target objects. Contrary to $\mathcal{J}_{background}$, we utilize LPIPS to quantify the difference between the two manipulated images. Similar to $\mathcal{J}_{detect}$, we divide $\mathcal{J}_{background}$ and $\mathcal{J}_{afterimage}$ using their respective valid areas. 
\begin{align}
    \mathcal{J}&_{afterimage} = \notag \\ 
    &\frac{LPIPS ( \mathcal{R} (  I \odot m_i )  \odot L, \mathcal{R} (  I \odot (\mathbbm{1}_{I} - L )) \odot L)}{ \mathcal{A} (  L ) } .    
    \label{eq:afterimage}
\end{align}

The overall score function of the judge module can be summarized as
\begin{align}
    \mathcal{J} =  \mathcal{J}_{background} + \lambda_{a} \mathcal{J}_{afterimage} + \lambda_{d} \mathcal{J}_{detect},
    \label{eq:overall}
\end{align}
where the hyper-parameters are empirically set as $\lambda_{a}$ = 90,000 and $\lambda_{d}$ = 0.5.

\textbf{Importance Map }
We define the importance of a pixel ($x$) as the expected judge score when the pixel is masked ($M(x) = 0$). The importance map ($\Phi$) of AURA is expressed as
\begin{align}
    \Phi_{I,\mathcal{R}} (x) = \E_{M} [ \mathcal{J} (\mathcal{R}(I \odot M)) |\,  M(x) = 0 \ ], 
\end{align} 
where $ M : \Omega \mapsto \{ 0, 1 \}$ indicates a random mask with distribution $\mathcal{D}$.
The equation can be reformulated as a summation over mask $m$:
\begin{align}
    \Phi_{I,\mathcal{R}} (x) &= \sum_{m}^{} \mathcal{J} (\mathcal{R}(I \odot m)) P[M = m | M(x) = 0] \notag \\ 
    & = \sum_{m}^{} \mathcal{J} (\mathcal{R}(I \odot m)) \frac{P[M = m , M(x) = 0]}{P[ M(x) = 0]}. 
\label{equatioin1}
\end{align} 
As
\begin{align}
     P[M=m, M(x) = 0] 
    & = \twopartdef { 0, } {m(x) = 1} {P[M=m], } {m(x)=0} \notag \\
    & = (1 - m(x)) P[M = m],  
\end{align}
Eq. \ref{equatioin1} can be written as
\begin{align}
     \Phi_{I,\mathcal{R}} (x) 
     = \sum_{m}^{} \mathcal{J} (\mathcal{R}(I \odot m)) \cdot (1 - m(x)) \cdot \frac{P[M=m]}{P[ M(x) = 0]}. 
\label{equation2}
\end{align} 
Eq. \ref{equation2} can be expressed in matrix form as:
\begin{align}
     \Phi_{I,\mathcal{R}} 
     = \sum_{m}^{} \mathcal{J} (\mathcal{R}(I \odot m)) \cdot (\mathbbm{1}_{I} - m) \cdot \frac{P[M=m]}{\mathbbm{1}_{I} - \E[M]}, 
\end{align} 
where $P[M(x) = 0] = 1 - P[M(x) = 1] =  1 - \E[M(x)]$. $\mathbbm{1}_{I}$ is a binary mask with the same size as the image $I$ and all pixel values equal to one.  

Similar to \cite{petsiuk2018rise}, we estimate the summation using Monte Carlo sampling. $N$ random masks are utilized to calculate the summation as follows:
\begin{align}
     \Phi_{I,\mathcal{R}} & \approx \sum_{i=1}^{N} \mathcal{J} (\mathcal{R}(I \odot m_i)) \cdot m_i \cdot \frac{1}{N \cdot (\mathbbm{1}_{I} - \E[M])} \notag \\
    & = \frac{1}{ N \cdot \mathbbm{1}_{I} -  \sum_{j=1}^{N} m_j} \sum_{i=1}^{N} \mathcal{J} (\mathcal{R}(I \odot m_i)) \cdot m_i.
\end{align}

\textbf{Candidate mask generation }
AURA generates candidate masks $C$ using predefined percentiles $P = \{x \in \mathbb{N} | 1\leq x \leq p\}$. 
Let $\mathcal{T}$ be a function that finds the $100 - P_j$-th percentile score ($\phi_j$) of an importance map as follows:
\begin{align}
    \phi_j = \mathcal{T} ( \Phi, 100 - P_j).
\end{align}
As AURA figures out which pixels need to be further masked in the target segmentation mask, the percentage of the segmented area of the target segmentation mask is added to the percentiles.
\begin{align}
    \phi_j = \mathcal{T} \left( \Phi, 100 - P_j - \frac{\mathcal{A}(L)}{H \cdot W} \cdot 100 \right)
\end{align}
Then, the $j$-th candidate mask $C_j$ is obtained by masking pixels whose importance is greater or equal to the percentile score ($\phi_j$). 
\begin{align}
    C_j(x) = \twopartdef { 1 } {\Phi(x) \geq \phi_j} {0} {\Phi(x) < \phi_j}
\end{align}
Finally, AURA selects the output mask ($C^*$) with the highest score evaluated by the judge module among the candidate masks: 
\begin{align}
    C^{*} = \mathop{\arg \max}\limits_{C_{j}} \mathcal{J}(\mathcal{R}(I \odot C_j)),
\end{align}
where $j \in P$.

\begin{algorithm}
\caption{Automatic mask generation using randomized input sampling}
\textbf{Input}: \\ image $I$, target segmentation map $L$ \\
\textbf{Output}: AURA mask $C^*$ \\
Sample $m_{\{1, \ldots ,N\}}$ \\

\For{i = 1 : N}{$ \Phi \gets \Phi + \mathcal{J}(\mathcal{R}(I \odot m_i)) \cdot (\mathbbm{1}_{I} - m_{i}) $\\
$m_{sum} \gets m_{sum} + ( \mathbbm{1}_{I} - m_{i})$}

$\Phi \gets \Phi \oslash m_{sum}$ \\
$P \gets \{1, \ldots , p \} $ \\
$H,W \gets shape(I)$ \\

\For{j = 1 : len(P)}{
$\phi_{j} \gets \mathcal{T} \left( \Phi , 100 - P_{j} - \frac{\mathcal{A}(L)}{H*W}\right)$ \\
$C_j \gets \Phi > \phi_{j}$
}

$C^{*} \gets \mathop{\arg \max}\limits_{C_{j}} \mathcal{J}(\mathcal{R}(I \odot C_j)) $ \\
\label{algo:AURA}
\end{algorithm}

\begin{table*}[]
\centering

\begin{tabular}{cc|cccc|cccc}
\hline
\multicolumn{2}{c|}{Mask} & \multicolumn{4}{c|}{Car} & \multicolumn{4}{c}{Human} \\ \hline
\multicolumn{1}{c|}{\multirow{6}{*}{Segmentation mask}} & kernel\_size & FID $\downarrow$ & LPIPS $\downarrow$ & PSNR $\uparrow$ & SSIM $\uparrow$ & FID $\downarrow$ & LPIPS $\downarrow$ & PSNR $\uparrow$ & SSIM $\uparrow$ \\ \cline{2-10} 
\multicolumn{1}{c|}{} & 0 & \underline{65.78} & 0.080 & \underline{32.42} & \underline{0.884} & 47.06 & 0.077 & \underline{\textbf{31.64}} & \underline{0.875} \\
\multicolumn{1}{c|}{} & 10 & 65.84 & \underline{0.072} & 32.38 & 0.880 & \underline{46.62} & \underline{0.075} & 31.53 & 0.859 \\
\multicolumn{1}{c|}{} & 20 & 68.87 & 0.073 & 32.36 & 0.869 & 57.24 & 0.091 & 31.35 & 0.829 \\
\multicolumn{1}{c|}{} & 30 & 76.46 & 0.078 & 32.29 & 0.851 & 67.42 & 0.110 & 31.16 & 0.799 \\
\multicolumn{1}{c|}{} & 40 & 82.43 & 0.084 & 32.20 & 0.832 & 76.25 & 0.134 & 30.97 & 0.768 \\ \hline
\multicolumn{2}{c|}{AURA mask (Proposed)} & \underline{\textbf{61.17}} & \underline{\textbf{0.069}} & \underline{\textbf{32.43}} & \underline{\textbf{0.887}} & \underline{\textbf{45.37}} & \underline{\textbf{0.073}} & \underline{31.62} & \underline{\textbf{0.877}} \\ \hline
\end{tabular}

\caption{
Object removal performance of target segmentation masks and AURA masks on the RORD dataset. We report FID, LPIPS, PSNR, and SSIM. The best and second-best results are underlined with bold letters and underlined, respectively. \cite{suvorov2022resolution} is utilized as an object remover. 
}
\label{table:RORD}
\end{table*}

\section{Experiments and results}
\textbf{Dataset and evaluation method }
We conduct object removal experiments on the COCO \cite{lin2014microsoft} and RORD \cite{Sagong_2022_BMVC} datasets. Similar to \cite{cao2021learning}, we use images whose target segmentation masks cover 5--40\% of the images. The validatation set images of COCO and RORD are utilized for the experiments. As RORD provides an image with no moving objects in a scene and images with various moving objects in the identical scene, we can utilize images without moving objects as the ground truth (GT) for object removal. Therefore, we assess the quality of object removal with image quality assessment (IQA) methods that rely on reference images for evaluation, such as PSNR, SSIM, and LPIPS. We also use FID calculated using the object removal GT for the comparison. 

The COCO dataset does not provide object removal GT. Therefore, for quality comparison, we utilize FID$^*$ and U-IDS$^*$ \cite{10536158}, designed to gauge the object removal quality without using the object removal GT images. While FID and U-IDS do not impose any condition within the query and comparison sets, FID$^*$ and U-IDS$^*$ are calculated using class-wise object removal results as a query set and images without a target class object as a comparison set. To leverage these evaluation methods, we designate all target class objects in the image as removal targets and use images without target class objects as the comparison set. 

\textbf{Implementation details }
The proposed method utilizes 2000 samples per image. We use Alexnet \cite{krizhevsky2014one} to calculate LPIPS. The image inpainting network \cite{suvorov2022resolution} trained on the PLACE \cite{zhou2017places} is employed as an object remover. We use 20 as the largest percentile ($p$).

\begin{algorithm}
\caption{Random mask sampling}
\textbf{Input}: number of masks N, \\ target segmentation map $L$ \\
\textbf{Output}: random mask $M$ \\
$H,W \gets shape(I)$ \\

\For{i = 1 : N}{$N_{patches} = random.uniform(3,5)$ \\
$M_{i} \gets (\mathbbm{1}_{M} - L)$  \\
    \For{$j = 1:N_{patches}$}
    {\uIf{$j<4$}{
    $ (x,y) \gets where(L==1) $\\
    }
    \Else{
    $ x \gets random.uniform(-r, W+r) $\\
    $ y \gets random.uniform(-r, H+r) $\\
    }
    $r \gets random.uniform(10,40)$ \\
    $ x \gets random.uniform(x-r, x+r) $\\
    $ y \gets random.uniform(y-r, y+r) $\\
    $ M_i \gets opencv.circle(M_i, (x,y), r)$} }    
\label{algo:mask}
\end{algorithm}

\textbf{Mask generation }
We generate random masks by adding circular patches to a target segmentation mask as in Algorithm \ref{algo:mask}. To generate a random mask, we first sample the number of additional patches from a uniform distribution between 3 and 5. Then, the radius is sampled from the uniform distribution between [10,40].

\subsection{Quantitative evaluation}
We compare the quality of object removal results made by AURA masks with those obtained using target segmentation masks and enlarged masks as shown in Table \ref{table:RORD}. The quality of the object removal results obtained using AURA masks is rated the highest when evaluating the quality using FID and LPIPS, which calculate image quality using high-level features extracted by a pre-trained CNN. Evaluations of SSIM and PSNR on object removal results obtained from the segmentation maps demonstrate that the quality decreases as the size of the input mask increases. In evaluations using SSIM and PSNR, even though AURA masks use wider areas than target segmentation masks using $kernel\_size=0$, the quality of object removal results obtained using AURA masks is evaluated higher than those using the target segmentation masks in three out of four evaluations. 

\begin{table}[t!]

\centering
\caption{PAR evaluation on the RORD car removal results.}

\begin{tabular}{c|ccccc}
\hline
Kernel size & 0 & 10 & 20 & 30 & 40 \\ \hline
PAR & 3.656 & 2.657 & 2.310 & 2.570 & 2.297 \\ \hline
\end{tabular}
\label{table:PAR}

\end{table}

\cite{zhang2022perceptual} introduces PAR, a non-reference inpainted image quality assessment method. The PAR evaluations of the car removal results on the RORD dataset are presented in Table \ref{table:PAR}. PAR evaluates that the quality of object removal results increases as the size of the input mask increases, which does not align with any evaluations made by the image quality assessment methods (FID, LPIPS, PSNR, and SSIM). Therefore, we do not utilize PAR for quality evaluation. 

In Table \ref{table:RORD}, all evaluation methods rate the quality of object removal results generated by target segmentation masks using $kernel\_size=0$ or $10$ as the best among those obtained using target segmentation masks and their variants. Therefore, for the experiment using the COCO dataset, we consider only the segmentation masks using $kernel\_size=0$ and $10$ as the baseline. 

The experimental results conducted on the COCO dataset are presented in Table \ref{table:coco} and \ref{table:u-ids}. Unlike the RORD dataset, the COCO dataset does not provide object removal ground truth (GT). Therefore, quality evaluation is conducted using FID$^*$ and U-IDS$^*$ \cite{10536158}, designed to assess object removal quality in the absence of object removal GT. We also utilize human evaluators for the quality evaluation. 

Table \ref{table:coco} shows that FID$^*$ evaluates the quality of object removal results obtained using the AURA masks as the best. As shown in Table \ref{table:u-ids}, U-IDS$^*$ also judges that using AURA masks as input masks results in the highest quality object removal. U-IDS$^*$ utilizes a support vector machine for the quality evaluation, which can easily distinguish inpainted (fake) samples from real samples if the number of total samples is small. Therefore, only object removal results of $person$ class (more than 12K samples) can be properly evaluated using the U-IDS$^*$. Through these results, we confirm that we can obtain better object removal results on the COCO images by using AURA masks instead of masks generated from segmentation maps.

\begin{table*}[]
\begin{tabular}{cc|cccccccccc|c}
\hline
\multicolumn{2}{c|}{Mask} & \multirow{2}{*}{airplane} & \multirow{2}{*}{cat} & \multirow{2}{*}{dog} & \multirow{2}{*}{fire hydrant} & \multirow{2}{*}{giraffe} & \multirow{2}{*}{horse} & \multirow{2}{*}{person} & \multirow{2}{*}{sheep} & \multirow{2}{*}{teddy bear} & \multirow{2}{*}{zebra} & \multirow{2}{*}{Votes} \\ \cline{1-2}
\multicolumn{1}{c|}{\multirow{3}{*}{\begin{tabular}[c]{@{}c@{}}Segmentation \\ mask\end{tabular}}} & kernel\_size &  &  &  &  &  &  &  &  &  &  &  \\ \cline{2-13} 
\multicolumn{1}{c|}{} & 0 & 170.63 & 99.34 & 102.03 & 187.56 & 158.06 & 123.56 & 50.72 & 187.23 & 144.24 & 218.59 & 10 \\
\multicolumn{1}{c|}{} & 10 & 168.70 & 91.35 & 93.94 & 184.65 & 166.03 & 120.45 & 45.63 & 191.48 & 134.25 & 218.40 & 72 \\ \hline
\multicolumn{2}{c|}{AURA mask (proposed)} & \underline{\textbf{162.58}} & \underline{\textbf{90.29}} & \underline{\textbf{92.46}} & \underline{\textbf{182.73}} & \underline{\textbf{156.96}} & \underline{\textbf{119.38}} & \underline{\textbf{44.60}} & \underline{\textbf{185.02}} & \underline{\textbf{133.14}} & \underline{\textbf{217.94}} & \underline{\textbf{158}} \\ \hline
\end{tabular}
\caption{
Object removal performance of target segmentation masks and AURA masks on the COCO dataset. We report FID$^*$. We remove the target class objects in the images of the COCO validation set. Images without each target class objects are utilized as a comparison set. The best results are underlined with bold letters. \cite{suvorov2022resolution} is utilized as an object remover. 
}
\label{table:coco}
\end{table*}

\begin{table}[]
\centering
\begin{tabular}{c|cc|c}
\hline
Comparison set                                                           & \begin{tabular}[c]{@{}c@{}}Seg. mask\\ kernel size= 0\end{tabular} & \begin{tabular}[c]{@{}c@{}}Seg. mask\\ kernel size = 10\end{tabular} & \begin{tabular}[c]{@{}c@{}}AURA mask\\ (proposed)\end{tabular} \\ \hline
\begin{tabular}[c]{@{}c@{}}Train set\\ w/o person\end{tabular} & 0.1167                                                             & 0.1295                                                               & \textbf{0.1301}                                                \\ \hline
\begin{tabular}[c]{@{}c@{}}Val. set\\ w/o person\end{tabular}  & 0.1205                                                             & 0.1325                                                               & \textbf{0.1348}                                                \\ \hline
\end{tabular}
\caption{
U-IDS$^*$ results on the COCO dataset. $person$ class objects are removed in the images of the COCO validation set. We use the COCO validation and train set images without the $person$ class objects as the comparison set, respectively. 
}
\label{table:u-ids}
\end{table}

We conduct a user study to acquire human judgments on which mask best removes objects in an image. We ask 20 users to find the best one of the three images. We make three completed images using AURA masks, target segmentation masks, and enlarged target segmentation masks using $kernel\_size = 10$. A user makes 12 decisions, and the results are presented in the last column of Table \ref{table:coco}. The human evaluators judge that the object removal results using AURA masks have the best quality, which is consistent with the evaluation made by FID$^*$ and U-IDS$^*$. In other words, three different evaluation methods make the identical assessment that the AURA mask can better remove the target class objects than a target segmentation mask and its variants. 

\begin{figure*}[th!]
\centering
\includegraphics[width=0.80\linewidth]{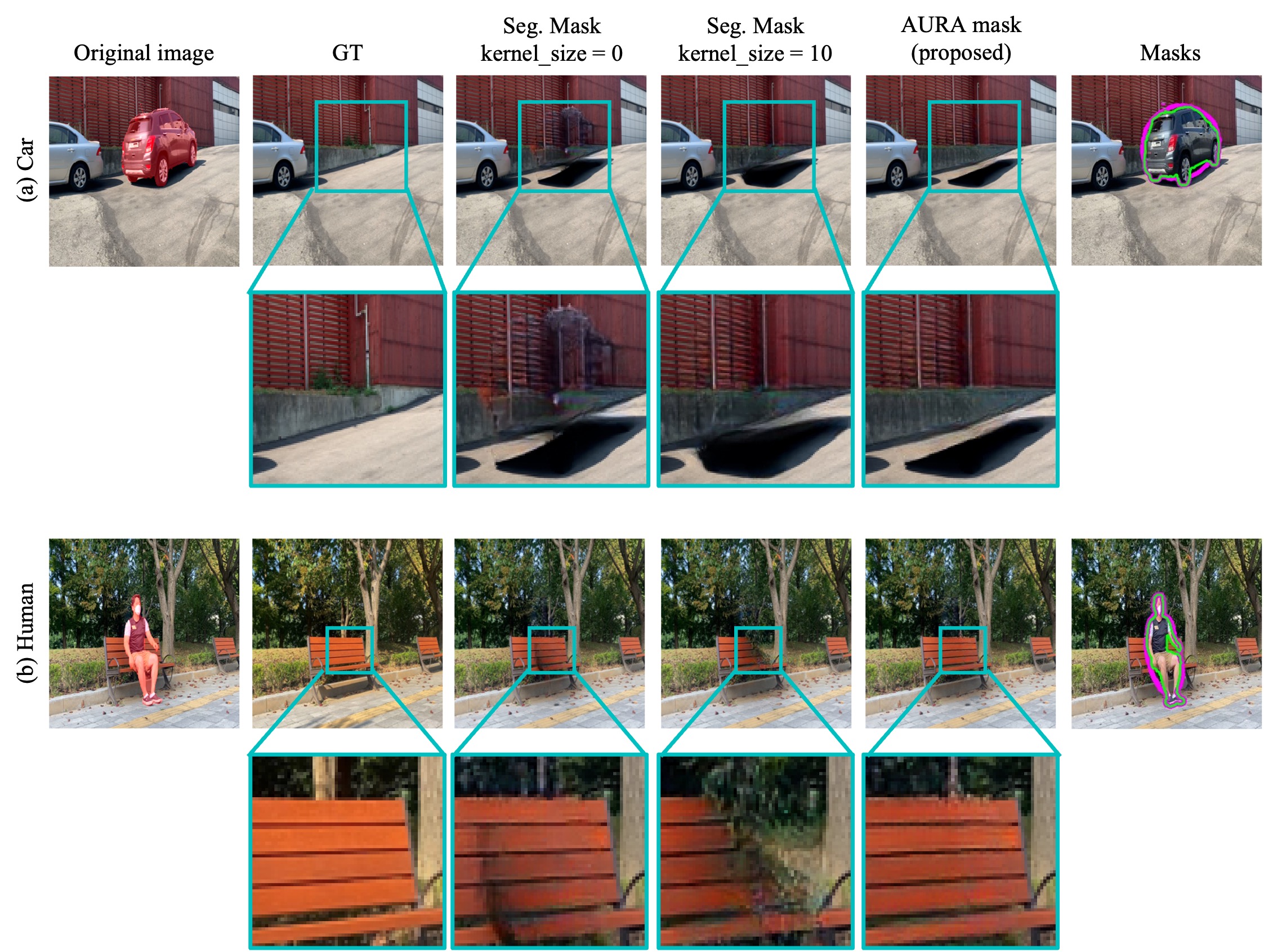} 
\caption{Object removal results on the RORD dataset images. Green and magenta lines indicate target segmentation masks and AURA masks, respectively. }  
\label{fig:qual_rord}
\end{figure*}

\begin{figure*}[th!]
\centering
\includegraphics[width=1.0\linewidth]{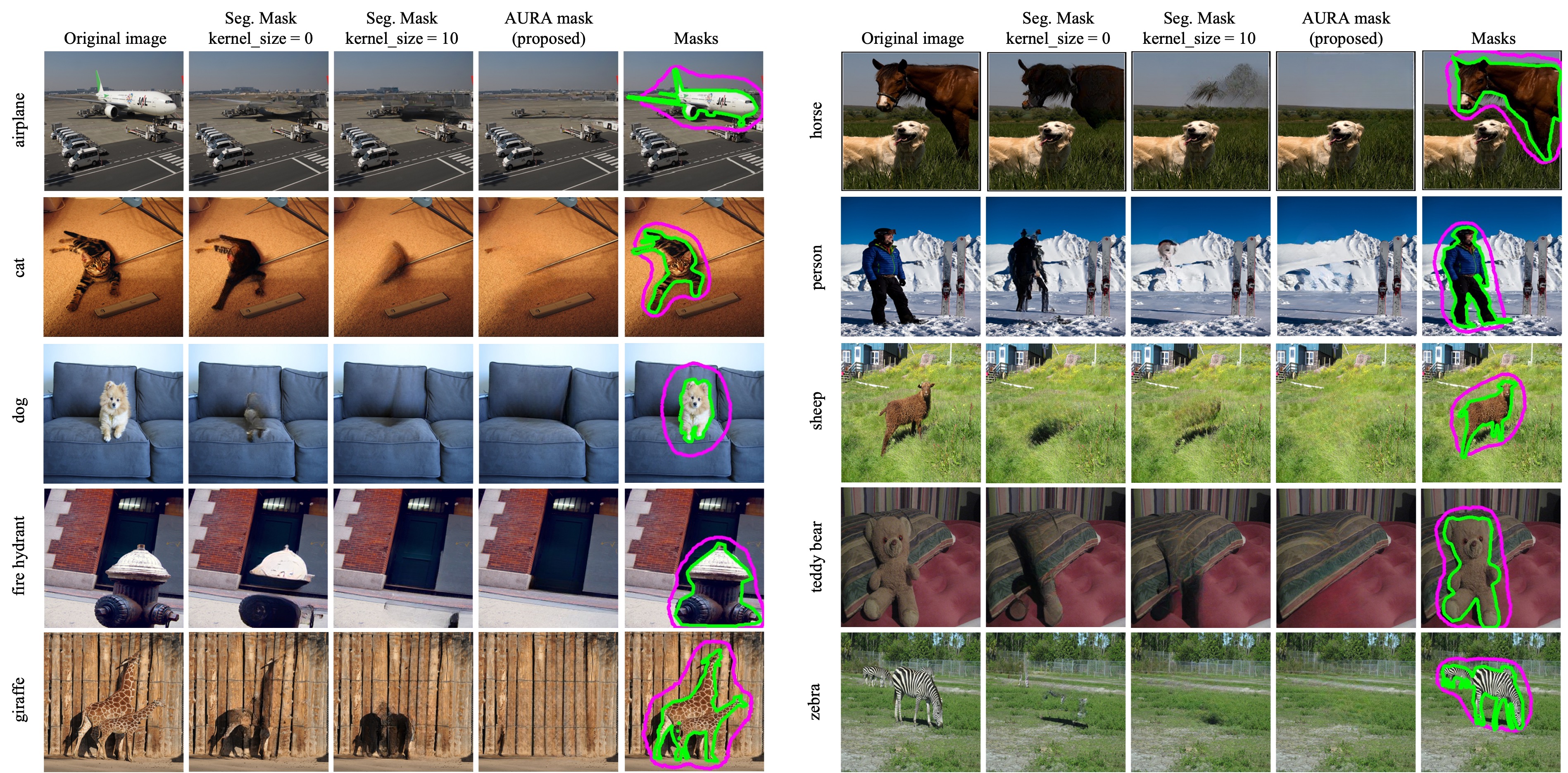} 
\caption{Object removal results on the COCO dataset images. Green and magenta lines indicate target segmentation masks and AURA masks, respectively. }  
\label{fig:qualitative}
\end{figure*}

\subsection{Qualitative evaluation}
Figs. \ref{fig:qual_rord} and \ref{fig:qualitative} show the qualitative results. We can confirm that AURA masks remove target objects in the most visually plausible way. We can easily observe the afterimage of the removal targets (Fig. \ref{fig:qual_rord} (a) and Fig. \ref{fig:qualitative}) or unsatisfactory restoration around removal targets (Fig. \ref{fig:qual_rord} (b)) in the object removal results obtained using the segmentation maps. These results demonstrate that to obtain high-quality object removal results without using the proposed method, a user needs to check whether any afterimages remain or if the surrounding background is excessively damaged. In other words, human intervention is required to judge which object removal result is the best among the results obtained using a semantic segmentation network. In contrast, the proposed method can automatically generate input masks that can make good object removal results without human intervention. 

\begin{figure}[t!]
\centering
\includegraphics[width=0.8\linewidth]{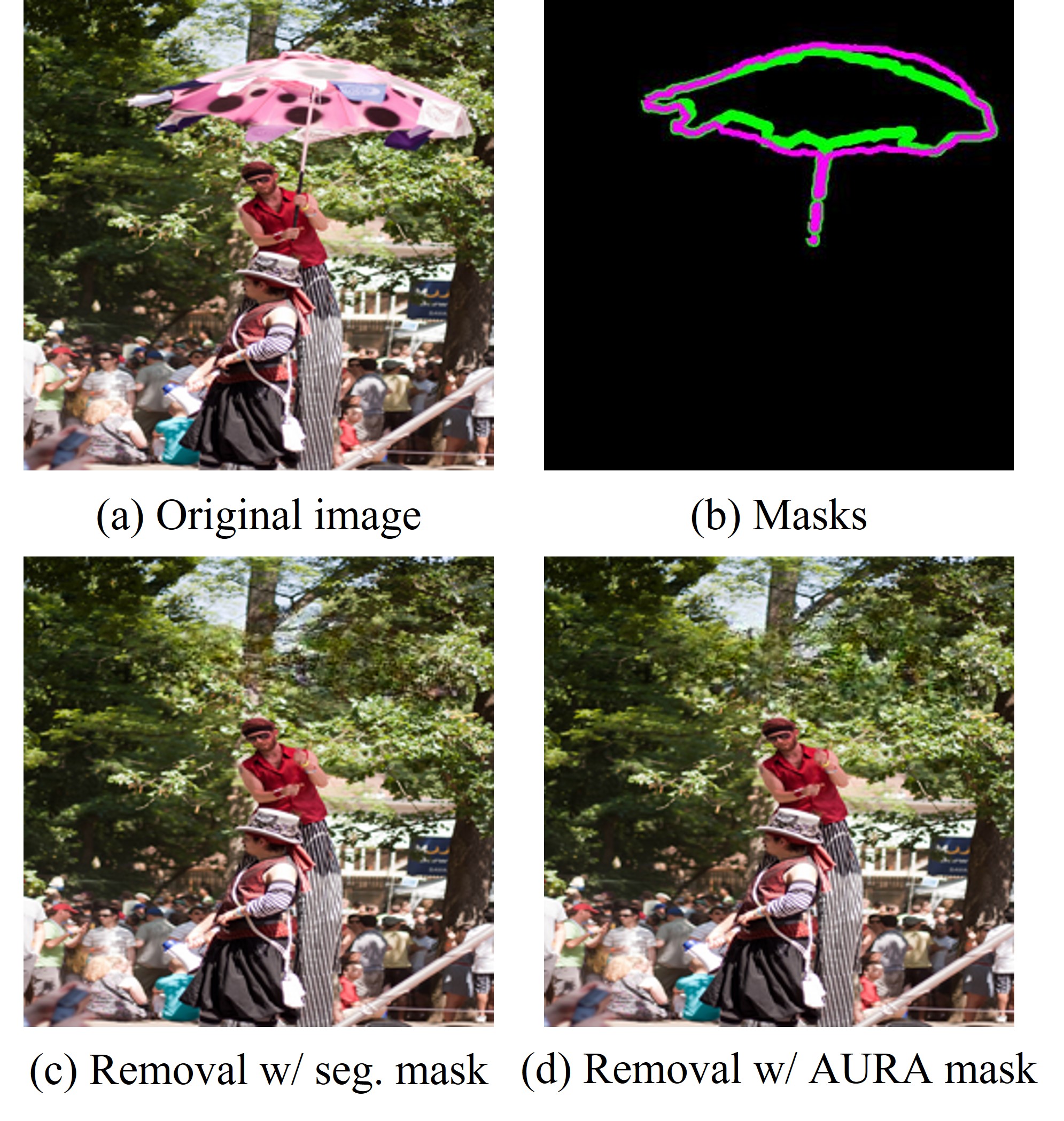} 
\caption{Object removal using the target segmentation mask (green) and the AURA mask (magenta). The umbrella is set as the removal target.}
\label{supp_fig1}
\end{figure}

Fig. \ref{supp_fig1} (c) shows a case where a target segmentation mask erases removal targets in a plausible way. As shown in Fig. \ref{supp_fig1} (d), the proposed method can generate a mask that erases the removal target well, even in this scenario. Recall that $\mathcal{J}_{afterimage}$ tries to enlarge the masked area so that the regions completed using a query mask become more dissimilar to the corresponding regions using a target segmentation mask. Therefore, when a target segmentation mask yields good inpainting, $\mathcal{J}_{afterimage}$ encourages the completed regions using a query mask to be different from the existing good result. Thus, $\mathcal{J}_{afterimage}$ induces AURA to generate a worse result. However, contrary to $\mathcal{J}_{afterimage}$, $\mathcal{J}_{background}$ encourages AURA to use as few additional masked areas as possible, which prevents AURA from generating a worse result.  And it should be noted that $\mathcal{J}_{background}$ acts more on the judge score than $\mathcal{J}_{afterimage}$ when a target segmentation mask removes objects well. This is because additional masking makes no significant perceptual differences in the regions of interest when a target segmentation mask already makes good inpainting, as shown in Fig. \ref{supp_fig1} (c) and (d). Therefore, AURA utilizes a small number of pixels for additional masking and generates a mask similar to a target segmentation mask when the segmentation mask removes objects well. The proposed method further masks only 3\% of the pixels to generate the AURA mask in the example (Fig. \ref{supp_fig1}. (b)).

\subsection{Number of query masks}
We believe that for object removal tasks, users are often willing to accept extended processing times in exchange for high-quality results. Thus, our method utilizes multiple removal processes to obtain better quality outcomes. We check the minimum number of samples needed to reliably obtain high-quality object removal results using the proposed method. Table \ref{sample_N_table} shows the FID and LPIPS of the AURA masks obtained by setting the number of samples differently. Table \ref{sample_N_table} demonstrates that the object removal results obtained using AURA masks generated from 2K or more samples have the same LPIPS value as those obtained using AURA masks generated from 5K samples. And we confirm that the average FID obtained using the AURA masks generated from 2K samples is within the standard deviation of the average FID value obtained using 5K samples for AURA mask generation. These results demonstrate that although the mask generation process contains random processes, we can obtain decent object removal results when using 2K samples for mask generation. 

\begin{table*}[t!]
\centering
\begin{tabular}{c|cccccc}
\hline
N   & 500    & 1000    & 2000   & 3000   & 4000 & 5000  \\ \hline
FID $\downarrow$& $64.58 \pm 0.34$ & $62.82 \pm 0.30$ & $61.17 \pm 0.12$ & $61.09 \pm 0.11$ & $61.08 \pm 0.10$ & $61.00 \pm 0.12$ \\ 
LPIPS $\downarrow$& $0.072 \pm 0.000$ & $0.071 \pm 0.000$ & $4.23 \pm 0.069$ & $0.069 \pm 0.000$ & $0.069 \pm 0.000$ & $0.069 \pm 0.000$ \\ \hline
\end{tabular}
\caption{FID and LPIPS of the AURA masks obtained by setting the number of random samples (N) differently. We report the average values and standard deviations for 5 independent runs. \cite{suvorov2022resolution} is utilized for inpainting. We use RORD car removal results for evaluation.}
\label{sample_N_table}
\end{table*}

\begin{table}[t!]
\centering
\begin{tabular}{c|cc|cc}
\hline
 & \multicolumn{2}{c|}{FID $\downarrow$} & \multicolumn{2}{c}{LPIPS $\downarrow$} \\ \hline
$\mathcal{J}_{afterimage}$ using L2 & 63.06 & (+1.89) & 0.072 & (+0.003) \\
w/o $\mathcal{J}_{detect}$ & 61.46 & (+0.29) & 0.069 & (+0.00) \\
w/o $\mathcal{J}_{background}$ & 84.86 & (+23.69) & 0.100 & (+0.031) \\
w/o $\mathcal{J}_{afterimage}$ & 63.48 & (+2.31) & 0.073 & (+0.004) \\ \hline
Proposed & 61.17 & - & 0.069 & - \\ \hline
\end{tabular}
\caption{
Ablation study on the Judge module. We report FID and LPIPS evaluations on the RORD car removal results. \cite{suvorov2022resolution} is utilized for inpainting.
}
\label{table:ablation}
\end{table}

\subsection{Ablation Study}
\textbf{Judge Module }
Ablation studies on the judge module are presented in Table \ref{table:ablation}. We observe that the absence of either $\mathcal{J}_{afterimage}$ or $\mathcal{J}_{background}$ increases FID and LPIPS. $\mathcal{J}_{background}$ induces the mask to cover the object tightly, while $\mathcal{J}_{afterimage}$ induces the mask to be large enough to cover the object loosely. Therefore, when either one of the two components is missing, the objective of AURA is no longer generating a mask that removes objects well. Therefore, the quality of the completed images decreases. The third row of Table \ref{table:ablation} shows that $\mathcal{J}_{detect}$ also helps improving the quality of the completed images. The result indicates that reducing the cases where the silhouette of the afterimage looks like the objects' outline helps to improve the quality of the completed images.

\textbf{Loss function }
The judge module evaluates object removal results by quantifying 1) how much the after image of the target class objects remains and 2) whether the backgrounds have been well restored.  We use L2 loss to calculate the difference between the original image’s background region and the completed image’s corresponding region. However, to compare the regions of removal targets in the completed images using the target segmentation mask and query mask, we utilize LPIPS \cite{zhang2018unreasonable} to calculate the difference between the two regions. We empirically confirm that the AURA mask using LPIPS to calculate $J_{afterimage}$ can better remove the target objects than the AURA mask using L2 loss to calculate $J_{afterimage}$, as shown in the third row of Table \ref{table:ablation}. When the surrounding pixels of the target objects have low RGB values and pixels from the completed regions have low RGB values, $J_{afterimage}$ using L2 loss may give a low score even if the restoration is well done. This is because L2 loss cannot make huge difference between the pixels from the well-restored regions and pixels from the afterimage. However, in the case of using LPIPS, $J_{afterimage}$ can give a high score to the well-restored regions with low RGB values, as the metric estimates the difference using the high-level image feature representations of the completed regions. For this reason, we infer that the proposed method can make better masks using LPIPS to calculate $J_{afterimage}$.

\section{Conclusion}
We propose an automatic mask generation method to better remove objects using an image inpainting network. The proposed method calculates the judge scores of randomly sampled masks and generates an importance map based on the randomly sampled masks and their judge scores. Among the candidate masks obtained using an importance map, the proposed method selects the mask with the highest judge score as the AURA mask. Experiments on the large image datasets demonstrate that AURA masks can better remove objects than target segmentation masks and enlarged target segmentation masks.

{\small
\bibliographystyle{IEEEtran}
\bibliography{bib_list}
}

\vfill

\end{document}